\documentclass[10pt,twocolumn,letterpaper]{article}

\usepackage{cvpr}              %
\usepackage{siunitx}

\usepackage{graphicx}
\usepackage{amsmath}
\usepackage{amssymb}
\usepackage{booktabs}
\usepackage{amsfonts}
\usepackage[utf8]{inputenc}
\usepackage{algorithm}
\usepackage{algpseudocode}
\usepackage{bbding}
\usepackage{color}
\usepackage{multirow}
\usepackage{makecell}
\usepackage{algpseudocode}
\usepackage{caption}   %
\usepackage{longtable} %
\usepackage{array}     %

\usepackage{caption}
\definecolor{cvprblue}{rgb}{0.21,0.49,0.74}
\usepackage[pagebackref,breaklinks,colorlinks,allcolors=cvprblue]{hyperref}

\def\paperID{12941} %
\def\confName{CVPR}
\def\confYear{2026}

\title{\raisebox{-0.5em}{\includegraphics[height=1.8em]{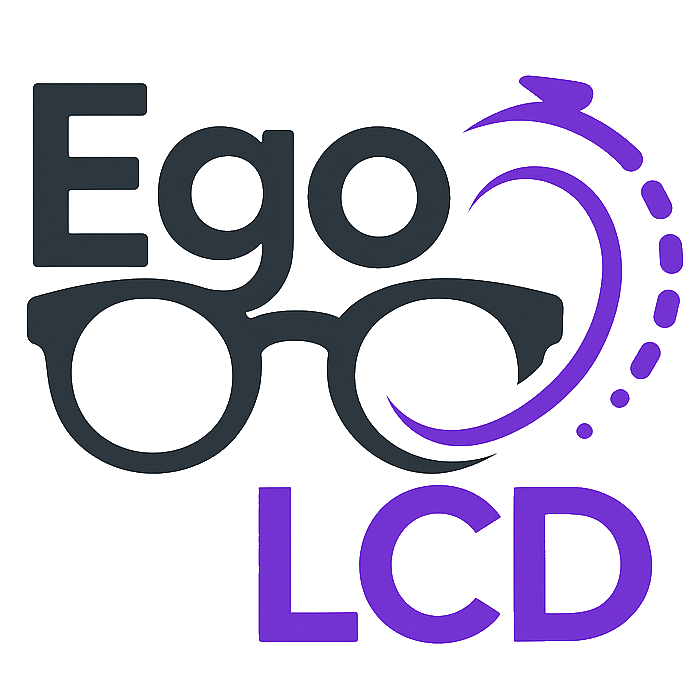}}~EgoLCD: Egocentric Video Generation with Long Context Diffusion}

\author{Liuzhou Zhang$^{1*}$~~
Jiarui Ye$^{1*}$~~
Yuanlei Wang$^{2*}$~~
Ming Zhong$^{3}$~~
Mingju Gao$^{4}$~~
Wanke Xia$^{5}$\\
Bowen Zeng$^{3}$~~
Zeyu Zhang$^{1\dag}$~~
Hao Tang$^{1\ddag}$\\
\normalsize $^{1}$Peking University~~
$^{2}$Sun Yat-sen University~~
$^{3}$Zhejiang University~~
$^{4}$Chinese Academy of Sciences~~
$^{5}$Tsinghua University\\
\footnotesize $^*$Equal contribution. $^\dag$Project lead.
$^\ddag$Corresponding authors: bjdxtanghao@gmail.com.}

\begin{document}

\twocolumn[{
\maketitle
\centering
\centering
\resizebox{\linewidth}{!}{
\includegraphics{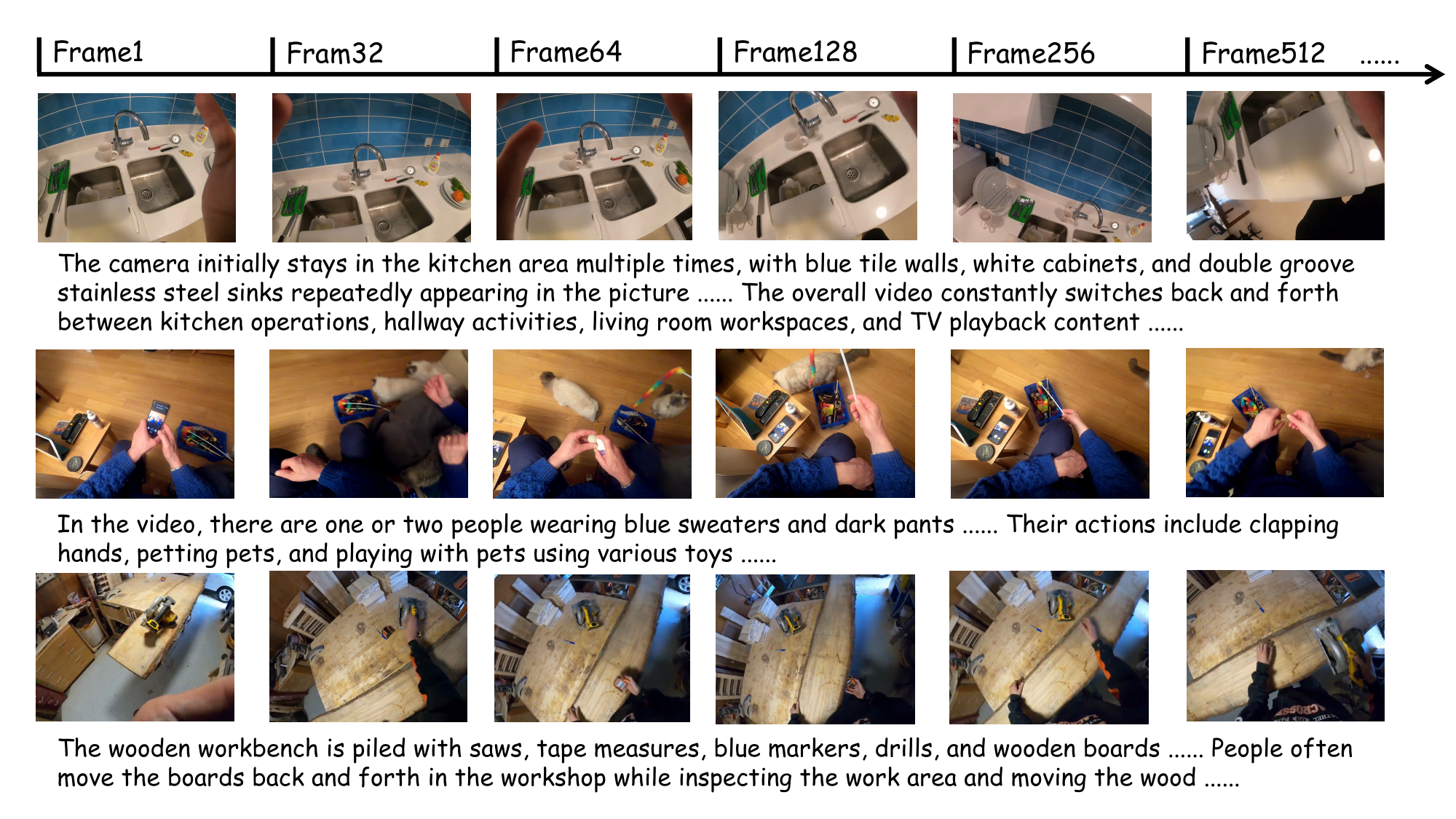}}
\captionof{figure}{The porposed EgoLCD generates long-form egocentric videos that maintain coherent scene transitions and consistent object layouts.}
\label{fig:fig0}
\vspace{0.4cm}
}]

\begin{abstract}

Generating long, coherent egocentric videos is difficult, as hand-object interactions and procedural tasks require reliable long-term memory. Existing autoregressive models suffer from content drift, where object identity and scene semantics degrade over time. To address this challenge, we introduce EgoLCD, an end-to-end framework for egocentric long-context video generation that treats long video synthesis as a problem of efficient and stable memory management. EgoLCD combines a Long-Term Sparse KV Cache for stable global context with an attention-based short-term memory, extended by LoRA for local adaptation. A Memory Regulation Loss enforces consistent memory usage, and Structured Narrative Prompting provides explicit temporal guidance. Extensive experiments on the EgoVid-5M benchmark demonstrate that EgoLCD achieves state-of-the-art performance in both perceptual quality and temporal consistency, effectively mitigating generative forgetting and representing a significant step toward building scalable world models for embodied AI. 
Code: \url{https://github.com/AIGeeksGroup/EgoLCD}.
Website: \url{https://aigeeksgroup.github.io/EgoLCD}.

\end{abstract}

\section{Introduction}

The paradigm of video generation~\cite{shi2025presentagent,wang2025drivegen3d,liu2025fpsattention,luo2025univid} is rapidly evolving beyond content creation towards the ambitious goal of building learned ``world simulators" \cite{sora2024, bruce2024genie, hu2023gaia1}. Such models, capable of simulating future outcomes in response to actions, hold immense potential for training and evaluating embodied AI agents in rich, diverse environments that transcend the limitations of manually engineered simulators \cite{hafner2023dreamerv3, ha2018world}. A critical yet underexplored frontier in this domain is the simulation of the human experience from a first-person perspective. Egocentric video, which captures a wearer's direct interactions with the world, offers an unparalleled data source for teaching agents complex, multi-step tasks involving fine-grained object manipulation \cite{shah2024egomimic, zhang2025egovla, nagarajan2021shaping}. Large-scale egocentric datasets like Ego4D and EPIC-KITCHENS provide a window into this experience, but leveraging them to build useful simulators requires generating long, logically coherent, and computationally tractable video sequences \cite{grauman2022ego4d, damen2020epic}.

However, generating long videos remains a formidable challenge, as it fundamentally strains a model's computational resources and its capacity for long-term memory~\cite{zhang2025blockvid,team2025inferix}. The quadratic complexity of self-attention in standard Transformers makes processing long sequences computationally prohibitive \cite{huang2025moc}. Autoregressive (AR) models, which maintain a generative memory via Key-Value (KV) caching to condition future frames on the past, offer an efficient alternative \cite{lu2025ca2vdm, zhou2025causvid}. Yet, this generative memory is often fragile. AR models are notoriously prone to ``content drift"—a form of generative forgetting where the model gradually loses track of object identity, appearance, and scene semantics over time. This issue is particularly acute for egocentric video, where high-variance camera motion, complex hand-object dynamics, and the necessity for strict procedural logic (e.g., following a recipe) mean that even minor memory lapses can render a simulation useless for agent training.

To address these challenges, we introduce EgoLCD (Egocentric Long Context Diffusion), an end-to-end framework that treats long video generation as a problem of robust and efficient memory management. Our approach integrates innovations from low-level memory operators to high-level reasoning paradigms. We also recognize that existing metrics inadequately capture the temporal nature of content drift. Standard evaluations often average quality scores, failing to penalize models whose memory degrades quickly. This motivates our development of a new evaluation protocol to better assess long-term temporal stability. Through this multi-faceted approach, EgoLCD generates long-form egocentric videos, as showcased in Fig.\ref{fig:fig0}, that achieve a new standard in consistency and logical coherence, marking a significant step towards building robust, scalable world models for embodied AI~\cite{song2025maniplvm,song2025hazards,huang20253d,liu2025nav,ye2025vla,huang2025mobilevla,liu2025evovla,huang20253dr1,huang2025dc,wu2025stereoadapter}.

In summary, our main contributions are:
\begin{itemize}

\item The proposed \textbf{EgoLCD} is built on a long-context diffusion model with a \textit{long–short memory} design for egocentric long-video generation. Specifically, the \textit{Long-Term Sparse KV Cache} serves as a long-term memory, caching historical key–value pairs and retrieving them based on their importance to the target video clip. Meanwhile, the attention mechanism with a limited context window acts as the core short-term memory, dynamically enhanced by LoRA parameters to enable fast adaptation to recent visual contexts during egocentric video generation. From the training perspective, we address the issue of forgetting in long video generation by introducing a \textit{memory regulation loss} that maintains alignment between historical memory and newly learned representations.

\item A novel \textit{Structured Narrative Prompting} (SNP) methodology for training and inference. By providing a temporally-ordered sequence of detailed captions, SNP serves as an external memory script, guiding the model through complex action sequences and enhancing its generation of precise temporal details, aligning with recent work on structured video captioning \cite{fan2025instancecap, wu2025any2caption}.

\item To comprehensively validate the superiority of EgoLCD in long-form egocentric video generation, we conducted extensive experiments on the authoritative EgoVid-5M benchmark\cite{wang2024egovid}, building upon our two-stage training framework utilizing both general video corpora and egocentric datasets. We recognize that existing metrics, by averaging quality scores, fail to effectively capture content drift in long sequences. This motivated us to develop a novel evaluation protocol specifically designed to assess long-term temporal stability. Experimental results demonstrate that EgoLCD, refined through two-stage training, not only leads in all key metrics on the EgoVid benchmark but also exhibits exceptional temporal consistency under our new protocol, systematically proving its potential as a high-performance egocentric video world model.

\end{itemize}
\begin{figure*}[t]
\centering
\resizebox{1\linewidth}{!}{
\includegraphics{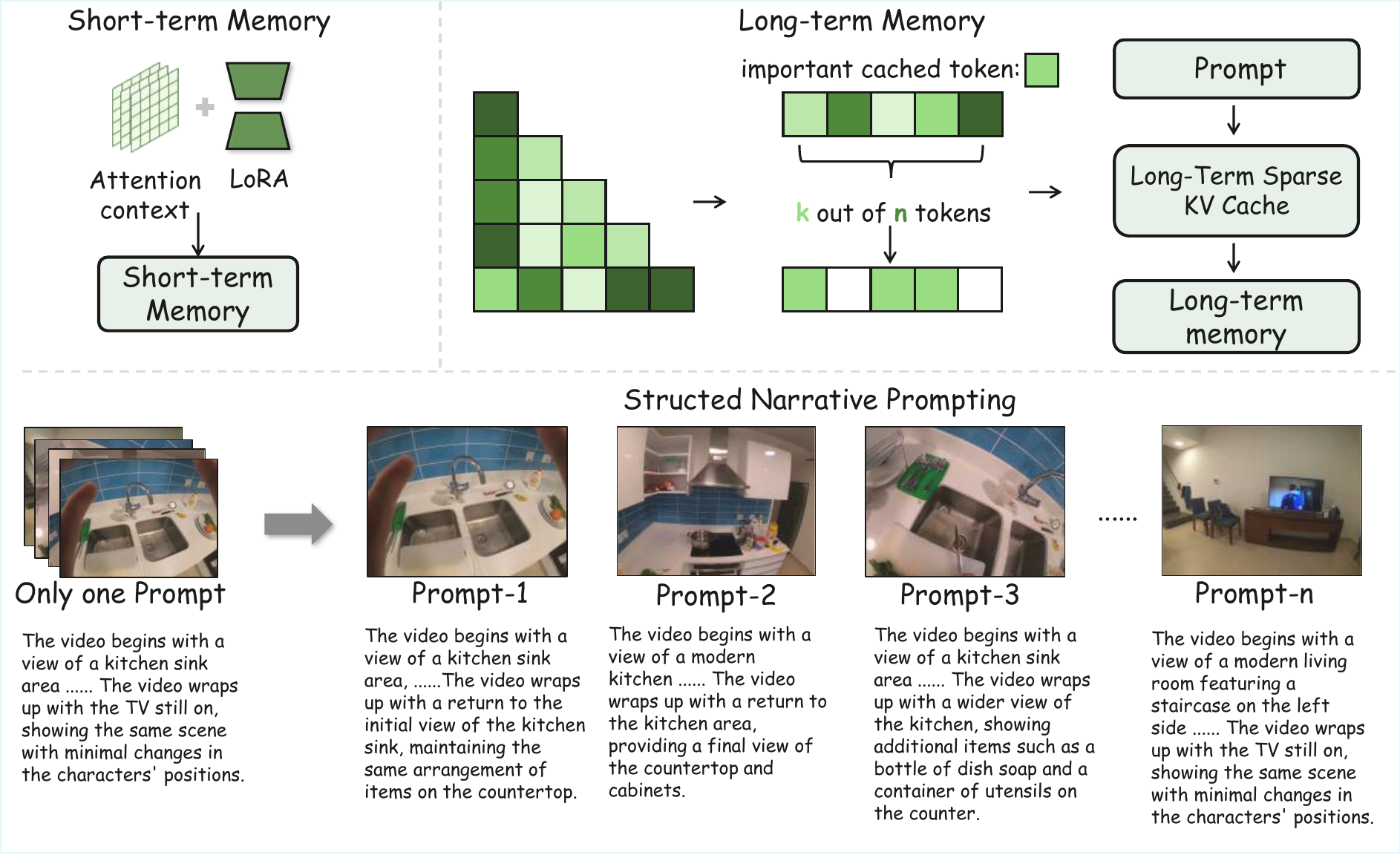
}}
\caption{Long-Short Memory \& Structured Narrative Prompting.}
\label{fig:LSM-SNP}
\end{figure*}

\section{Related Work}

\subsection{Video Generation as World Simulators}
The ``World Model" concept has evolved from model-based reinforcement learning to large-scale generative models. Initially, Ha and Schmidhuber used it to improve RL sample efficiency via a compressed latent space \cite{ha2018world}, and models like DreamerV3 advanced it by learning behaviors in imagined trajectories \cite{hafner2023dreamerv3}. Recent models like Sora, GAIA-1, and Genie extend this idea to high-fidelity simulators trained on large video datasets \cite{sora2024,  brooks2024genie}. This has led to a distinction between models for \textit{internal world understanding} and \textit{external world simulation}. Benchmarks like WorldSimBench and EVA-Bench now assess visual quality and physical plausibility \cite{qin2024worldsimbench, li2024eva}.

While driving simulators model macro-level interactions, egocentric video datasets like Ego4D and EPIC-Kitchens \cite{grauman2022ego4d, damen2018epic} capture fine-grained human-object interactions. Works like EgoMimic and EgoVLA show that policies trained on first-person human and robot data improve manipulation \cite{shah2024egomimic, zhang2025egovla}. EgoLCD extends this, proposing that long, coherent egocentric video generation is key for simulating complex human tasks.

\subsection{Long Video Generation}
Long video generation is constrained by the quadratic complexity of self-attention in Transformers, leading to the adoption of autoregressive (AR) models with Key-Value (KV) caching \cite{lu2025ca2vdm, zhou2025causvid}. Models like Ca2-VDM and CausVid optimize this pipeline by sharing caches and compressing the denoising process \cite{lu2025ca2vdm, zhou2025causvid}. Techniques like sub-2-bit quantization and dynamic layer-wise allocation further optimize KV caching \cite{liu2025vidkv, wan2025meda}. A major issue in AR models is "content drift," where visual and semantic identity degrade over time. Solutions include motion-guided losses, optical flow propagation, and temporal correlation structuring \cite{zhang2024mgld, zhou2024upscale, chen2024artdiff}. Some models blend low-frequency features for continuity with high-frequency features for detail \cite{wang2025freelong}.

To address content drift, recent models focus on long-range dependency modeling, such as Mixture of Contexts (MoC) \cite{huang2025moc} and Long Context Tuning (LCT) \cite{li2025lct}. Our work, EgoLCD, contributes by developing a diffusion model for long-context egocentric video generation.

\section{The Proposed Method}
\label{sec:method}
\begin{figure*}[t]
\centering
\resizebox{\linewidth}{!}{
\includegraphics{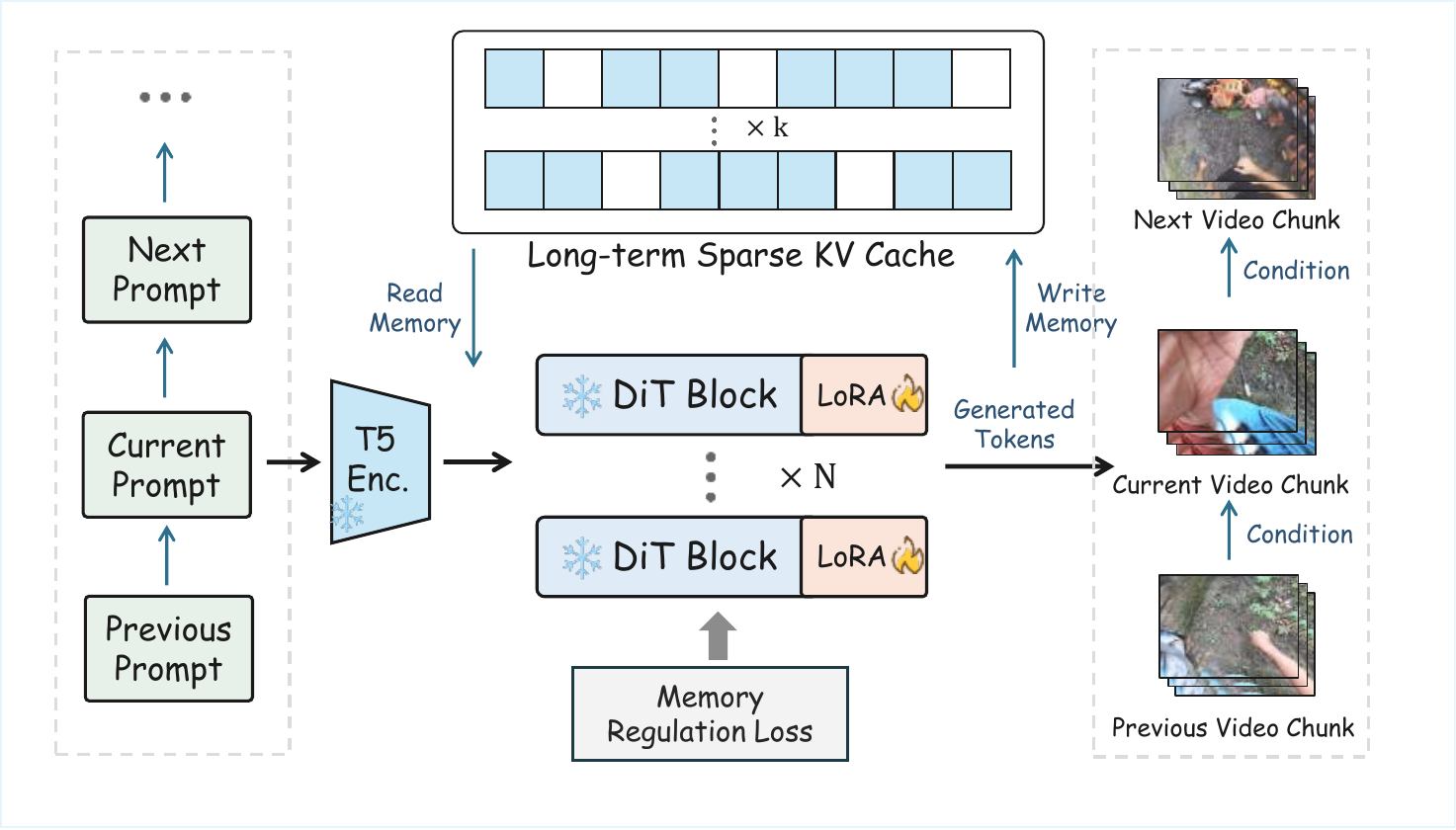}}
\caption{The overall framework of \textit{EgoLCD}.}
\label{fig:egolcd}
\end{figure*}

\subsection{Overview}
\label{subsec:overview}
As shown in Fig.~\ref{fig:egolcd}, EgoLCD is a long-context diffusion framework for egocentric world modeling that generates temporally coherent long video sequences through a block-wise architecture. It produces video segments sequentially, with each new segment conditioned on the preceding context to maintain long-range consistency.

The framework employs a semi-autoregressive (Semi-AR) diffusion strategy for scalable long video synthesis. Current methods face several limitations: (1) inefficient context propagation causes temporal drift; (2) memory requirements grow quadratically with sequence length; (3) training-inference inconsistency in memory management; (4) difficulty balancing long-term coherence with short-term adaptation to dynamic egocentric views.

EgoLCD addresses these through a novel long-short-term memory design, featuring a long-term sparse KV cache for global context stability and a short-term memory module for local pattern adaptation. The short-term memory rapidly captures evolving scene dynamics while selectively consolidating key information into long-term storage.

EgoLCD enhances generation quality through structured narrative prompting and memory regulation loss. The narrative system maintains continuity by storing and retrieving prompts from a semantic cache, while the memory loss constrains generation using historical segments as semantic anchors. These complementary approaches jointly ensure semantic and temporal consistency in long video generation.

\subsection{Long-Short Memory}
\label{subsec:memory}
Inspired by Titans~\cite{behrouz2024titans}, where attention acts as short-term memory and a neural memory module captures persistent history, we adopt a dual-memory design for EgoLCD. In our framework, the \textit{Long-Term Sparse KV Cache} functions as a long-term memory for distant dependencies, while the attention mechanism with a limited context window serves as short-term memory, enhanced by LoRA parameters as implicit memory units, as shown in Fig.~\ref{fig:LSM-SNP}. This separation allows us to both retain long-horizon consistency and support rapid local adaptation—mitigating forgetting in long egocentric video generation, aided by the integration of a \textit{Memory Regulation Loss} to ensure consistency in long-term memory usage.

\paragraph{Sparse KV Caching as Long-Term Memory.} 
Maintaining a semantically consistent memory is essential for generating high-quality long content for egocentric video generation. However, storing all memory as KV Cache is very expensive. Hence, we propose a \textit{Long-Term Sparse KV Cache} that bridges the training-inference gap through unified memory operations. Our approach employs consistent sparse caching strategies across both phases, enabling efficient long-term dependency modeling while eliminating performance degradation for sustainable first-person video generation.

Let $\mathcal{H}$ denote the memory repository storing historical prompt embeddings and their corresponding sparse KV caches. For each new prompt embedding $E_t$, we compute semantic relevance to historical embeddings $E_u$ via cosine similarity:
\begin{equation}
s_u = \frac{\langle E_t, E_u \rangle}{\|E_t\| \|E_u\|}, \quad \forall E_u \in \mathcal{H},
\end{equation}
where $s_u$ represents the semantic similarity score, $E_t$ denotes the current prompt embedding, and $E_u$ indicates the historical prompt embeddings in the memory repository.

The retrieval set $\mathcal{R}_t$ containing the top-$m$ most relevant historical prompts enables contextual fusion through concatenation operations:
\begin{equation}
\tilde{K}_t = \text{Concat}(K_t, K_{\mathcal{R}_t}), \quad \tilde{V}_t = \text{Concat}(V_t, V_{\mathcal{R}_t}),
\end{equation}
where $\tilde{K}_t$ and $\tilde{V}_t$ represent the fused key and value tensors, $K_t$ and $V_t$ denote the current key-value pairs, and $K_{\mathcal{R}_t}$ and $V_{\mathcal{R}_t}$ correspond to the retrieved historical key-value pairs.

The sparse caching mechanism employs probe-based importance scoring for memory compression. Given the probe token set $\mathcal{P}$, we compute attention weights:
\begin{equation}
A^{(l)}_{\text{probe}} = \text{Softmax}\left(\frac{Q_{\mathcal{P}}K^T}{\sqrt{d_k}} + M\right),
\end{equation}
where $A^{(l)}_{\text{probe}}$ denotes the attention weight matrix for layer $l$, $Q_{\mathcal{P}}$ represents probe queries, $K$ indicates all keys, $d_k$ is the key dimension, and $M$ signifies the causal mask matrix.

The importance score $\sigma_p^{(l)}$ for each token $p$ aggregates attention weights across probes and heads:
\begin{equation}
\sigma_p^{(l)} = \sum_{i=1}^{|\mathcal{P}|}\sum_{h=1}^{H^{(l)}} A^{(l)}_{i,p,h},
\end{equation}
where $|\mathcal{P}|$ denotes the probe set size and $H^{(l)}$ represents the number of attention heads in layer $l$.

Position-aware normalized importance accounts for temporal locality in egocentric sequences:
\begin{equation}
\tilde{\sigma}_p^{(l)} = \frac{\sigma_p^{(l)}}{L - p + 1},
\end{equation}
where $\tilde{\sigma}_p^{(l)}$ indicates the normalized importance score, $L$ signifies the sequence length, and $p$ denotes the token position.

The sparse token set $\mathcal{I}^{(l)}$ preserves tokens with maximal cumulative importance:
\begin{equation}
\begin{aligned}
r^{(l)} & = \min\left\{ m \geq 1 \mid \sum_{u=1}^{m} \sigma_{\pi^{(l)}(u)}^{(l)} \geq \tau \sum_{v=1}^{L} \sigma_v^{(l)} \right\}, \\
\mathcal{I}^{(l)} & = \{\pi^{(l)}(1), \pi^{(l)}(2), \ldots, \pi^{(l)}(r^{(l)})\},
\end{aligned}
\end{equation}
where $r^{(l)}$ represents the number of retained tokens, $\tau$ denotes the importance threshold, and $\pi^{(l)}$ indicates the importance sorting permutation.

During generation, each query $\mathbf{q}_t$ attends exclusively to the sparse cache:
\begin{equation}
\begin{aligned}
\mathbf{h}_t^{(l)} &= \sum_{j \in \mathcal{I}^{(l)}} \alpha_{t,j}^{(l)} V_j^{(l)},\\
\alpha_{t,j}^{(l)} & = \frac{\exp\left(\langle \mathbf{q}_t^{(l)}, K_j^{(l)} \rangle / \sqrt{d_k}\right)}{\sum_{k \in \mathcal{I}^{(l)}} \exp\left(\langle \mathbf{q}_t^{(l)}, K_k^{(l)} \rangle / \sqrt{d_k}\right)},
\end{aligned}
\end{equation}
where $\mathbf{h}_t^{(l)}$ denotes the output hidden state, $\alpha_{t,j}^{(l)}$ represents the attention weight, and $V_j^{(l)}$ indicates the sparse value cache.

This unified memory framework establishes persistent knowledge trajectories through $\mathcal{H}$, enables context-aware retrieval via $\mathcal{R}_t$, and achieves efficient compression through $\mathcal{I}^{(l)}$, providing an effective solution for egocentric long-video generation while maintaining sub-linear memory complexity and ensuring training-inference consistency.

\paragraph{Attention as Short-Term Memory with LoRA Enhancement.}
Drawing inspiration from recent advances in visual memory~\cite{ma2024visual}—which reinterpret lightweight adaptation layers such as LoRA as implicit memory units—we conceptualize the limited-context-window attention mechanism as the core short-term memory module within the EgoLCD framework while treating LoRA parameters as auxiliary implicit memory units. In egocentric video generation, the visual scene evolves continuously as the camera wearer moves, necessitating a mechanism capable of rapid adaptation to local changes without disrupting the long-term context. The attention mechanism, by leveraging its constrained context window and precise dependency modeling, functions as short-term memory, enabling the model to dynamically maintain transient representations within each generation block while preserving stable long-term information in the sparse KV cache. Concurrently, LoRA parameters serve as implicit memory units, providing lightweight and adaptive enhancements to the attention mechanism. The transient representations held in the attention context are progressively integrated into long-term memory via the sparse KV cache over time. This dual design empowers EgoLCD to effectively balance persistent world modeling with moment-to-moment egocentric variations, achieving agile short-term adaptation while upholding long-term coherence.

\paragraph{Memory Regulation Loss.}
To tightly couple the dual-memory design with the generative dynamics, we introduce a \textit{Memory Regulation Loss} that regularizes the predicted velocity field using semantically retrieved long-term context. Let \( x_t \in \mathbb{R}^{C \times T_0 \times H \times W} \) denote the ground-truth video clip at timestep \( t \), where \( C \), \( T_0 \), \( H \), and \( W \) represent the channel dimension, temporal length, height, and width, respectively, and let \( \epsilon_t \) be the sampled noise. The diffusion model predicts a velocity field \( v_\theta(x_t, t) \) following the rectified flow formulation, where \( \theta \) denotes the model parameters.

To inject long-range semantics into the supervision signal, we retrieve the top-\( m \) most relevant historical video segments from the memory repository \( \mathcal{H} \), denoted as \( \{ X^{(u)} \}_{u \in \mathcal{R}_t} \), where \( u \) indexes the retrieved segments and \( \mathcal{R}_t \) denotes the set of indices for the relevant historical segments. Since these segments may differ in temporal resolution, each segment is temporally resampled to match the current clip length \( T_0 \) as follows:
\begin{equation}
\hat{X}^{(u)} = \text{Resample}\big(X^{(u)}, T_0, H, W\big), \quad \forall u \in \mathcal{R}_t,
\end{equation}
where \( \hat{X}^{(u)} \) represents the resampled historical segment for the \( u \)-th segment, where \( X^{(u)} \) is the original segment. The resampled segments are then aggregated to form a semantic anchor representing the retrieved long-term context:
\begin{equation}
\bar{X}_{\text{cond}} = \frac{1}{|\mathcal{R}_t|} \sum_{u \in \mathcal{R}_t} \hat{X}^{(u)},
\end{equation}
where \( |\mathcal{R}_t| \) is the number of relevant historical segments retrieved from memory. \( \bar{X}_{\text{cond}} \) serves as the semantic anchor, a synthesized representation of the long-term context that aggregates information from multiple historical video segments.

Instead of supervising the velocity solely toward \( x_t \), we define a memory-regularized target velocity that incorporates this semantic anchor:
\begin{equation}
v^{*}_{\text{mem}} = \epsilon_t - \bar{X}_{\text{cond}},
\end{equation}
where \( \epsilon_t \) is the sampled noise, and \( v^{*}_{\text{mem}} \) is the target velocity conditioned on the memory, representing a refined velocity that aligns with the semantic anchor. We define the memory regulation loss as the expected squared error between the predicted velocity and the memory-regularized target velocity:
\begin{equation}
\mathcal{L}_{\text{mem}}
= \mathbb{E}\!\left[\left\| v_\theta(x_t, t) - v^{*}_{\text{mem}} \right\|_2^2 \right],
\end{equation}
where \( v_\theta(x_t, t) \) is the velocity predicted by the model for the ground-truth video at timestep \( t \).

A key advantage of this formulation is training–inference consistency. The model interacts with sparse memory retrieval during training in exactly the same way as in inference, eliminating the common train–test mismatch in long-horizon video generation. Additionally, by jointly optimizing retrieval, aggregation, and sparse memory fusion with the generative process, the memory module exhibits significantly improved stability and reliability during inference.

The final training objective integrates the proposed memory regulation loss with the rectified flow objective and the auxiliary reconstruction loss:
\begin{equation}
\mathcal{L}
= \mathcal{L}_{\text{RF}}
+ \lambda_{\text{MAE}} \mathcal{L}_{\text{MAE}}
+ \gamma \mathcal{L}_{\text{mem}},
\end{equation}
where \( \mathcal{L}_{\text{RF}} \) is the rectified flow loss, \( \mathcal{L}_{\text{MAE}} \) is the mean absolute error (MAE) loss, \( \lambda_{\text{MAE}} \) is a hyperparameter controlling the weight of the MAE loss, and \( \gamma \) is a hyperparameter controlling the weight of the memory regulation loss. This unified objective effectively stabilizes long-term memory usage, reduces drift across generation blocks, and significantly improves temporal coherence in egocentric long-video synthesis.

\subsection{Structured Narrative Prompting}
\label{sec:SNP}
To optimize the use of semantic caching and historical guidance, we propose a structured narrative prompting system for each 5-second video segment. The process begins by dividing the input video into smaller, 5-second clips, ensuring that each clip represents a coherent unit of action, scene, or transition. For each of these segments, GPT-4o~\cite{openai2024gpt4ocard} generates detailed captions that describe the visuals, characters, actions, and environment, capturing the essence of the scene.

During the inference process, the video is generated sequentially, with each 5-second segment being rolled out in succession. To maintain semantic consistency, we reference the prompts of previous segments. Specifically, we retrieve the most semantically similar prompts from earlier clips, leveraging a semantic KV cache to store these prompts along with their associated embeddings. These stored embeddings enrich the generation of the current segment by incorporating context from prior clips, ensuring smooth and coherent transitions both visually and narratively.

The structured narrative prompting serves two main purposes. First, it provides detailed guidance for generating each individual segment, ensuring that every 5-second clip contributes meaningfully to the overall narrative. Second, it forms the foundation for effectively utilizing the semantic KV cache. By storing and accessing relevant semantic information from previous clips, the system can dynamically adjust the generation of future segments, optimizing both visual consistency and narrative continuity throughout the entire video.

\section{Experiments}

\begin{table*}[t]
\centering
\small
\setlength{\tabcolsep}{4pt}
\sisetup{detect-all}
\begin{tabular}{l *{10}{S[table-format=1.4]}}
\toprule
& \multicolumn{5}{c}{\textbf{Perceptual quality metrics} ($\uparrow$)}
& \multicolumn{5}{c}{\textbf{NRDP metrics} ($\downarrow$)} \\
\cmidrule(lr){2-6} \cmidrule(lr){7-11}
\textbf{Method} &
{\makecell{\textbf{Img.}\\\textbf{Quality}}} &
{\makecell{\textbf{Motion}\\\textbf{Smooth.}}} &
{\makecell{\textbf{Aesth.}\\\textbf{Quality}}} &
{\makecell{\textbf{Bg.}\\\textbf{Consist.}}} &
{\makecell{\textbf{Subj.}\\\textbf{Consist.}}} &
{\textbf{Clarity}} &
{\textbf{Motion}} &
{\textbf{Aesthetic}} &
{\textbf{Background}} &
{\textbf{Subject}} \\
\midrule
MAGI \cite{ai2025magi1autoregressivevideogeneration}             & 0.6662 & 0.9947 & \textbf{0.6508} & \underline{0.9078} & 0.8992 & 2.7225 & \underline{0.0243} & 3.8286 & \underline{0.3090} & 0.6618 \\
Self-Forcing \cite{huang2025self}     & 0.6805 & 0.9947 & \underline{0.6283}          & \textit{0.8203} & 0.8481 & 3.0798 & 0.1549 & 3.4683 & 1.6108 & \underline{0.3716} \\
Framepack \cite{zhang2025framecontextpackingdrift}        & \textbf{0.6972} & \underline{0.9949} & 0.6043 & 0.8791 & \underline{0.9001} & 4.2513 & 0.0387 & \underline{1.4751} & 5.9421 & 4.3984 \\
SkyReels-v2 \cite{chen2025skyreels}           & 0.6567 & \textit{0.9926} & 0.5267 & 0.8924 & 0.8640 & 1.9503 & 0.0461 & 2.8957 & 0.9323 & 1.8292 \\
EgoLCD (Ours w/o loss) & 0.6643 & 0.9874 & \textit{0.5353} & 0.8895 & 0.8576 & \underline{1.8390} & 0.0509 & 2.0941 & 1.1124 & 1.7933 \\
EgoLCD (Ours)          & \underline{0.6852} & \textbf{0.9956} & 0.6047 & \textbf{0.9588} & \textbf{0.9597} & \textbf{0.7551} & \textbf{0.0119} & \textbf{0.9618} & \textbf{0.2945} & \textbf{0.0844} \\
\bottomrule
\end{tabular}
\caption{
Comparison of different methods on perceptual quality and NRDP-based no-reference metrics. 
The best scores in perceptual quality metrics are bolded and the second-best scores are italicized. 
In the table, \textbf{EgoLCD (w/o Loss)} refers to EgoLCD trained without the memory regulation loss.
}
\label{tab:full_metrics}
\end{table*}

\subsection{Datasets and Evaliation Metrics}
\label{sec:Datasets and Metrics}
\paragraph{Datasets.}
(1) \textbf{General Videos.} The first is a curated, general-domain dataset of 1,000 natural videos, which we collected to provide a foundation for learning general visual features and motion dynamics.
(2) \textbf{Egocentric Videos.} The second and primary source for domain specialization is the \textit{Ego4D} dataset. Ego4D is a massive-scale, diverse egocentric video dataset, offering 3,670 hours of daily-life activity videos that span hundreds of scenarios. This dataset is essential for our work as it provides extensive, long-form, first-person video that captures the unique viewpoints, camera motion, and complex hand-object interactions central to our task.

All video data from both sources were processed and converted into the Structured Narrative Prompting (SNP) format to serve as model inputs.

\paragraph{Evaliation Metrics.} Standard evaluation metrics for video generation, such as Fr\'{e}chet Video Distance (FVD), often rely on average scores across entire sequences. This approach is insufficient for long-context generation, as it obscures a critical failure mode: content drift. A model that generates a high-quality initial segment but degrades rapidly may achieve a misleadingly positive score.

To establish a comprehensive baseline, we first employ standard quality dimensions from VBench \cite{vbench}. VBench offers a fine-grained, hierarchical evaluation of video quality. We specifically utilize its objective metrics for \textit{Aesthetic Quality}, \textit{Imaging Quality}, \textit{Motion Smoothness}, \textit{Background Consistency} , and \textit{Subject Consistency} to assess the general performance of generated videos.

However, to specifically quantify temporal stability and penalize generative forgetting, we introduce a new evaluation metric: the \textit{Normalized Referenced Drifting Penalty (NRDP)}. The design of NRDP is guided by three key principles. First, it uses the initial video segment (the first chunk) as a high-quality reference, as error accumulation typically degrades subsequent chunks. Second, it applies a decaying penalty, imposing a stricter punishment on models that exhibit quality drift early in the sequence, which signals poorer core stability. Third, all drift calculations are normalized by the first chunk's quality, ensuring a fair comparison by removing biases from different models' baseline capabilities.
The NRDP calculation proceeds in four steps:
\begin{enumerate}
    \item The video is segmented into $N$ uniform chunks (we use $N=10$).
    \item A base quality metric $M$ (from VBench) is computed for each chunk, yielding a score sequence $M_1, \dots, M_N$.
    \item The normalized drift $D_i$ for each subsequent chunk $i$ is calculated as its relative deviation from the reference chunk $M_1$:
    \begin{equation}
    \label{eq:nrdp_drift}
    D_{i}=\frac{|M_{i}-M_{1}|}{M_{1}}.
    \end{equation}
    \item The final NRDP score is computed as a weighted sum of these drifts, using a monotonically decreasing weight $w_i$ (such as $w_i = N - i + 1$) to penalize early drift more severely:
    \begin{equation}
    \label{eq:nrdp_final}
    NRDP_{M}(V)=\sum_{i=2}^{N}w_{i}\cdot D_{i}.
    \end{equation}
\end{enumerate}

We apply NRDP to the VBench base metrics, resulting in our primary stability metrics: \textit{NRDP-Aesthetic}, \textit{NRDP-Motion}, and \textit{NRDP-Clarity} (derived from Imaging Quality).

\begin{table*}[t]
    \centering
    \small
  \resizebox{1\linewidth}{!}{
  \begin{tabular}{l|c|cccccc}
    \toprule
    \multicolumn{1}{l|}{Method} & w. \textit{EgoVid} & CD-FVD $\downarrow$ & Semantic Consistency $\uparrow$& Action Consistency $\uparrow$  & Clarity Score $\uparrow$ & Motion Smoothness $\uparrow$ & Motion Strength $\uparrow$\\
    \midrule
    SVD \cite{blattmann2023stable} & \XSolidBrush & 591.61 & 0.258 & 0.465 & 0.479 & 0.971 & 18.897\\
    SVD \cite{blattmann2023stable} & \Checkmark & 548.32 & 0.266 & 0.471 & 0.485 & 0.974 & 21.032 \\
    \midrule
    DynamiCrafter \cite{xing2024dynamicrafter} & \XSolidBrush & 243.63 &  0.257 & 0.481 & 0.473 & 0.986 & 9.357\\
    DynamiCrafter \cite{xing2024dynamicrafter} & \Checkmark & 236.82 & 0.265 & 0.494 & 0.483 & 0.987 & 18.329\\
    \midrule
    OpenSora \cite{zheng2024open} & \XSolidBrush & 809.46 & 0.260 & 0.489 & 0.520 &0.983 & 7.608 \\
    OpenSora \cite{zheng2024open} & \Checkmark & 718.32 & 0.266 & 0.494 & 0.528 & 0.986 & 15.871 \\
    \midrule
    EgoLCD (Ours) & \XSolidBrush & 187.94 & 0.291 & 0.510 & 0.530 & 0.992 & 20.732 \\
    EgoLCD (Ours) & \Checkmark & 177.23 & 0.298 & 0.517 & 0.540 & 0.995 & 22.351 \\
    \bottomrule
    \end{tabular}}
    \caption{Evaluations across six metrics confirm that training with the \textit{EgoVid} dataset leads to superior performance compared to the three baseline models.} 
    \label{tab:egovid-table}
\end{table*}

\subsection{Implementation Details}
EgoLCD is built upon the SkyReels-v2-1.3B latent diffusion transformer. We employ a two-stage training strategy: first on general videos for appearance and motion priors, and then on egocentric data for first-person specialization, with full memory components enabled.

For optimization, we use AdamW with a learning rate of 1e-5, a weight decay of 1e-4, and 200 warm-up steps. Training employs bf16 precision with FSDP (sequence-parallel degree 4), a global batch size of 8, and gradient checkpointing. The latent sequence length is capped at 75,600 tokens, with random dropout of video/KV conditions (0.2/0.1/0.1 probabilities) for robustness. A fixed negative prompt suppresses common artifacts.

We use a rectified-flow scheduler (1,000 steps) and optimize a hybrid loss that combines the rectified-flow objective, MAE reconstruction, and memory regulation loss. EMA decay is set to 0.99. Complete two-stage training requires approximately 50 hours on 8×H100 GPUs.

\subsection{Evaluation and Analysis on General Long Video Generation}

We evaluate EgoLCD against several state-of-the-art baseline models, including MAGI \cite{ai2025magi1autoregressivevideogeneration}, Self-Forcing \cite{huang2025self}, Framepack \cite{zhang2025framecontextpackingdrift}, and our base model, SkyReels-v2 \cite{chen2025skyreels}. The evaluation covers both standard perceptual quality metrics and our proposed NRDP metrics for measuring temporal drift, with results shown in Table \ref{tab:full_metrics}.

In terms of perceptual quality, EgoLCD achieves strong performance across multiple dimensions. It obtains the best scores in both Background Consistency and Subject Consistency, demonstrating the effectiveness of our long-short-term memory module in maintaining scene and object stability. The model also ranks first in Imaging Quality and Motion Smoothness, indicating good frame clarity and temporal coherence. While slightly lower than MAGI in Aesthetic Quality, EgoLCD shows meaningful improvement over its base model SkyReels-v2.

EgoLCD's most significant advantages appear in the NRDP metrics designed to measure content drift. Our method achieves the best performance across all five NRDP dimensions, substantially outperforming all baseline models. The notable reductions in NRDP-Subject and NRDP-Aesthetic confirm EgoLCD's ability to maintain initial segment quality throughout long generation sequences, effectively addressing generative forgetting. The ablation study (EgoLCD without loss) further validates the importance of Memory Regulation Loss, as its removal leads to a clear degradation in consistency performance.

\subsection{Performance Analysis in Egocentric Video Generation}

Table~\ref{tab:egovid-table} compares the propsoed EgoLCD with three baseline models: SVD \cite{blattmann2023stable}, DynamiCrafter \cite{xing2024dynamicrafter}, and OpenSora \cite{zheng2024open}, both with and without the \textit{EgoVid} dataset, specifically in the context of egocentric video generation. The results demonstrate that our two-stage training strategy is highly effective: by first pre-training on general videos to establish robust appearance and motion priors, and then specializing on egocentric data with full memory components enabled, EgoLCD achieves significant performance improvements across all six evaluation metrics when trained with \textit{EgoVid}. Notably, it achieves lower CD-FVD, indicating better temporal coherence, and shows improved semantic and action consistency, clarity, motion smoothness, and motion strength. Compared to the baseline models, EgoLCD outperforms them in all metrics, highlighting the effectiveness of its dual-memory design and the strategic use of the \textit{EgoVid} dataset in enhancing long-context consistency and dynamic motion for first-person perspective video generation.

\section{Limitations and Future Work}
Computationally, the sparse memory design still requires substantial GPU resources, and the fixed 5-second segment processing limits the temporal horizon. Methodologically, the reliance on precisely aligned text descriptions makes the model susceptible to noise-induced semantic drift, and the evaluation framework depends on automated metrics that may not reflect genuine user preferences.

 We plan to develop efficient memory management solutions to overcome generation length constraints and create noise-resistant narrative comprehension modules. Additionally, we will construct evaluation systems better aligned with human perception to reduce dependence on automated metrics.

\section{Conclusion}

This work presents EgoLCD, a framework addressing a key challenge in long video generation: maintaining visual and semantic consistency over extended sequences. Instead of treating egocentric synthesis as solely a generative task, EgoLCD reframes it as a memory management problem through a unified long–short memory design that preserves context while adapting to dynamic first-person scenes. The Long-Term Sparse KV Cache maintains global structure, while LoRA-enhanced short-term attention captures fine-grained motion and appearance changes. Memory Regulation Loss enforces alignment with historical context, and Structured Narrative Prompting provides temporal guidance across multi-stage actions. Evaluations on the EgoVid-5M benchmark show consistent gains in perceptual quality, semantic stability, and motion coherence, while substantially mitigating generative forgetting and temporal drift.

\clearpage
{
    \small
    \bibliographystyle{ieeenat_fullname}
    \bibliography{main}
}
\clearpage

\clearpage
\appendix
\setcounter{page}{1}
\maketitlesupplementary

\section{Algorithm Details}
\label{sec:appendix_algo}

In this section, we provide the detailed algorithmic procedures for the proposed EgoLCD framework, corresponding to the methodology described in Section~\ref{sec:method}.

\subsection{Inference Pipeline}
Algorithm~\ref{alg:inference} outlines the overall inference pipeline of EgoLCD. The process begins with \textbf{Structured Narrative Prompting} (Section~\ref{sec:SNP}), where the input script is segmented into coherent 5-second clips with detailed captions generated by GPT-4o. During the generation loop, the model utilizes \textbf{Semantic Retrieval} via cosine similarity (Section~\ref{subsec:memory}) to fetch relevant historical context from the database $\mathcal{H}$. This ensures that the current generation is conditioned on the most pertinent past events, effectively mitigating content drift in long-form video synthesis as discussed in Section~\ref{subsec:overview}.

\begin{algorithm}[htbp]
\caption{EgoLCD Inference Pipeline with Structured Narrative Prompting}
\label{alg:inference}
\begin{algorithmic}[1]
\Require Video Script $\mathcal{S}$, Pretrained DiT Model $\theta$, History Database $\mathcal{H} \leftarrow \emptyset$
\State \textbf{Preprocessing:} Split $\mathcal{S}$ into $N$ clips $\{C_1, \dots, C_N\}$ via GPT-4o
\For{$t = 1$ to $N$}
    \State Encode prompt $P_t$ to get embedding $E_t$
    
    \State \textbf{Memory Retrieval:}
    \If{$\mathcal{H}$ is not empty}
        \State Compute semantic relevance: $s_u \leftarrow \frac{\langle E_t, E_u \rangle}{\|E_t\| \|E_u\|}$
        \State Retrieve indices $\mathcal{R}_t \leftarrow \text{Top-}m(\{s_u\})$ based on relevance
        \State Fetch Historical KV: 
        \State \quad $K_{retr}, V_{retr} \leftarrow \text{Concat}(\{K^{(u)}, V^{(u)} \mid u \in \mathcal{R}_t\})$
    \Else
        \State $K_{retr}, V_{retr} \leftarrow \text{None}$
    \EndIf
    
    \State \textbf{Generation (Diffusion):}
    \State Initialize noise $z_t \sim \mathcal{N}(0, I)$
    \State Condition on previous clip $V_{t-1}$ (if $t>1$) to maintain continuity
    \For{diffusion step $s = T$ to $0$}
        \State $v_{pred} \leftarrow \text{Model}(z_t, s, P_t, K_{retr}, V_{retr})$ \Comment{See Alg.~\ref{alg:block}}
        \State $z_t \leftarrow \text{RectifiedFlowStep}(z_t, v_{pred})$
    \EndFor
    \State $V_t \leftarrow \text{Decode}(z_t)$
    
    \State \textbf{Memory Update:}
    \State $K_{sparse}, V_{sparse} \leftarrow \text{SparseCompress}(K_t, V_t)$ \Comment{See Alg.~\ref{alg:sparse}}
    \State Append $(E_t, K_{sparse}, V_{sparse})$ to $\mathcal{H}$
\EndFor
\State \Return Concatenated Video $V = [V_1, \dots, V_N]$
\end{algorithmic}
\end{algorithm}

\subsection{Dual-Memory DiT Block}
Algorithm~\ref{alg:block} details the internal mechanism of our modified DiT block, which realizes the \textbf{Long-Short Memory} design (Section~\ref{subsec:memory}). Within each block, the short-term memory is handled by the standard attention window, enhanced by LoRA parameters ($\phi$) to enable rapid adaptation to evolving egocentric viewpoints. Simultaneously, the long-term memory is injected by concatenating the retrieved sparse KV pairs ($K_{retr}, V_{retr}$) with the local features before the attention operation. This dual-pathway ensures the model maintains global consistency while responding to local dynamics.

\begin{algorithm}[htbp]
\caption{Dual-Memory DiT Block (Long-Short Context)}
\label{alg:block}
\begin{algorithmic}[1]
\Require Latent $x$, Current KV ($K_{cur}, V_{cur}$), Retrieved KV ($K_{retr}, V_{retr}$), LoRA params $\phi$
\State \textbf{Input Projection:}
\State $Q \leftarrow W_q x + \text{LoRA}_q(x)$ 
\State $K_{local} \leftarrow W_k x + \text{LoRA}_k(x)$
\State $V_{local} \leftarrow W_v x + \text{LoRA}_v(x)$

\State \textbf{Long-Short Fusion:}
\If{$K_{retr}$ is not None}
    \State $K_{fused} \leftarrow \text{Concat}([K_{retr}, K_{local}], \text{dim}=1)$
    \State $V_{fused} \leftarrow \text{Concat}([V_{retr}, V_{local}], \text{dim}=1)$
\Else
    \State $K_{fused}, V_{fused} \leftarrow K_{local}, V_{local}$
\EndIf

\State \textbf{Attention Mechanism:}
\State Compute attention weights with fused memory:
\State $A \leftarrow \text{Softmax}(\frac{Q K_{fused}^T}{\sqrt{d_k}})$ 
\State $O \leftarrow A V_{fused}$
\State $O \leftarrow W_o O + \text{LoRA}_o(O)$

\State \Return $O, K_{local}, V_{local}$
\end{algorithmic}
\end{algorithm}

\subsection{Sparse KV Compression}
To maintain computational efficiency and fixed memory overhead, we employ the \textbf{Sparse KV Compression} strategy described in Algorithm~\ref{alg:sparse}. Following the methodology in Section~\ref{subsec:memory}, we calculate the importance score $\sigma_p^{(l)}$ for each token by aggregating attention weights from probe tokens. These scores are normalized by position to account for temporal locality. Finally, a threshold-based pruning is applied to retain only the most critical key-value pairs for the Long-Term Sparse KV Cache.

\begin{algorithm}[htbp]
\caption{Sparse KV Compression Strategy}
\label{alg:sparse}
\begin{algorithmic}[1]
\Require Full KV Cache $K, V$, Sequence Length $L$, Probe Set $\mathcal{P}$
\State \textbf{Compute Importance:}
\For{layer $l$ in Transformer}
    \State Compute probe attention: $A_{probe}^{(l)} \leftarrow \text{Softmax}(\frac{Q_{\mathcal{P}} K^T}{\sqrt{d_k}} + M)$
    \State Aggregate importance: $\sigma_p^{(l)} \leftarrow \sum_{i \in \mathcal{P}} \sum_{h} A_{i,p,h}^{(l)}$
    \State Normalize position: $\tilde{\sigma}_p^{(l)} \leftarrow \frac{\sigma_p^{(l)}}{L - p + 1}$
\EndFor

\State \textbf{Token Pruning:}
\State Sort tokens by importance $\pi^{(l)}$
\State Determine cut-off $r^{(l)}$ where $\sum \tilde{\sigma} \ge \tau \sum \tilde{\sigma}_{total}$
\State Select indices $\mathcal{I}^{(l)} \leftarrow \{\pi^{(l)}(1), \dots, \pi^{(l)}(r^{(l)})\}$

\State \textbf{Compress:}
\State $K_{sparse} \leftarrow \text{Gather}(K, \mathcal{I}^{(l)})$
\State $V_{sparse} \leftarrow \text{Gather}(V, \mathcal{I}^{(l)})$

\State \Return $K_{sparse}, V_{sparse}$
\end{algorithmic}
\end{algorithm}

\section{Ablation Study}
\label{sec:ablation}

To validate the effectiveness of the core components in EgoLCD, we conduct an ablation study on the EgoVid-5M benchmark. It is important to note that all variants in this study, including the baseline, adopt the proposed block-wise autoregressive generation strategy and are conditioned on Structured Narrative Prompting (SNP). This ensures that the performance gains reported in Table~\ref{tab:ablation} stem strictly from our architectural contributions—the Long-Term Sparse KV Cache and the Memory Regulation Loss—rather than the inference framework or prompt engineering.

\paragraph{Impact of Long-Term Sparse KV Cache.}
The variant "EgoLCD (w/o KV Cache)" serves as our baseline (equivalent to the base model SkyReels-v2). Despite utilizing block-wise generation and detailed SNP guidance, this model suffers significantly from content drift, as evidenced by high error rates in NRDP-Subject (1.8292) and NRDP-Background (0.9323). This indicates that while SNP provides semantic continuity, it is insufficient for maintaining fine-grained visual consistency over long horizons without an explicit internal memory mechanism. By integrating the Long-Term Sparse KV Cache, the full EgoLCD model effectively bridges the gap between independent blocks, drastically reducing subject drift to 0.0844. This confirms that retrieving cached visual features is essential for preserving object identity across the block-wise generation process.

\paragraph{Impact of Memory Regulation Loss.}
The "EgoLCD (w/o loss)" variant incorporates the Sparse KV Cache but is trained without the Memory Regulation Loss ($\mathcal{L}_{mem}$). Although it outperforms the baseline by leveraging historical context, it still exhibits noticeable degradation compared to the full model (e.g., NRDP-Background 1.1124 vs. 0.2945). This suggests that simply attending to retrieved memory during the block-wise inference is not optimal; the model requires explicit supervision during training to learn how to align its generated velocity fields with the semantic anchors from the cache. The Memory Regulation Loss enforces this alignment, ensuring that the long-term context effectively regularizes the trajectory of each generation block.

\begin{table*}[t]
\centering
\small
\setlength{\tabcolsep}{4pt}
\sisetup{detect-all}
\begin{tabular}{l *{10}{S[table-format=1.4]}}
\toprule
& \multicolumn{5}{c}{\textbf{Perceptual quality metrics} ($\uparrow$)}
& \multicolumn{5}{c}{\textbf{NRDP metrics} ($\downarrow$)} \\
\cmidrule(lr){2-6} \cmidrule(lr){7-11}
\textbf{Method} &
{\makecell{\textbf{Img.}\\\textbf{Quality}}} &
{\makecell{\textbf{Motion}\\\textbf{Smooth.}}} &
{\makecell{\textbf{Aesth.}\\\textbf{Quality}}} &
{\makecell{\textbf{Bg.}\\\textbf{Consist.}}} &
{\makecell{\textbf{Subj.}\\\textbf{Consist.}}} &
{\textbf{Clarity}} &
{\textbf{Motion}} &
{\textbf{Aesthetic}} &
{\textbf{Background}} &
{\textbf{Subject}} \\
\midrule
EgoLCD (w/o KV Cache)           & 0.6567 & 0.9926 & 0.5267 & 0.8924 & 0.8640 & 1.9503 & 0.0461 & 2.8957 & 0.9323 & 1.8292 \\
EgoLCD (w/o loss) & 0.6643 & 0.9874 & 0.5353 & 0.8895 & 0.8576 & 1.8390 & 0.0509 & 2.0941 & 1.1124 & 1.7933 \\
EgoLCD            & 0.6852 & \textbf{0.9956} & \textbf{0.6047} & \textbf{0.9588} & \textbf{0.9597} & \textbf{0.7551} & \textbf{0.0119} & \textbf{0.9618} & \textbf{0.2945} & \textbf{0.0844} \\
\bottomrule
\end{tabular}
\caption{
Ablation study on the effectiveness of key components in EgoLCD. All methods utilize block-wise generation and Structured Narrative Prompting (SNP). "EgoLCD (w/o KV Cache)" denotes the baseline without the sparse memory module. "EgoLCD (w/o loss)" denotes the model with memory but trained without the Memory Regulation Loss. The results demonstrate that even with SNP and block-wise inference, the internal Sparse KV Cache and Regulation Loss are indispensable for minimizing temporal drift (NRDP metrics).
}
\label{tab:ablation}
\end{table*}

\section{Implementation Details}
\label{sec:implementation}

\paragraph{Training Configuration.}
We train the model on a cluster of 8 NVIDIA H100 GPUs using Full Sharded Data Parallel (FSDP) and bf16 precision. The optimization is performed using AdamW with a learning rate of $1 \times 10^{-5}$, a weight decay of $1 \times 10^{-4}$, and $\beta$ parameters implied by an epsilon of $1 \times 10^{-15}$. We apply a warmup of 200 steps and maintain an Exponential Moving Average (EMA) of model weights with a decay rate of 0.99. To enhance generation diversity, we employ classifier-free guidance training by randomly dropping video conditions ($p=0.2$) and text prompts ($p=0.1$). The training utilizes a dynamic bucket strategy for resolution management, with a base resolution of 480p, a batch size of 1 per GPU, and a sequence parallel degree of 1.

\paragraph{Inference Settings.}
For long-form video generation, we adapt the configuration to handle increased sequence lengths. We scale the sequence parallel degree to 4 to accommodate 81-frame sequences at a resolution of $480 \times 848$ (aspect ratio 9:16). The generation process uses the Rectified Flow scheduler with 20 sampling steps, a guidance scale of 5.0, and a sample shift of 5.0. To ensure temporal continuity across segments, we employ an autoregressive generation strategy where each new video segment is conditioned on the last 9 frames of the preceding segment. Furthermore, we explicitly cache and reuse the attention key-value states from previous steps to maintain long-term consistency and computational efficiency without re-computation. The maximum sequence length is capped at 75,600 tokens to fit within memory constraints.

\section{Structured Narrative Prompting (SNP) Details}
\label{sec:snp_details}

In this section, we provide concrete examples of our \textbf{Structured Narrative Prompting (SNP)} strategy. We employ distinct prompting structures for the training and inference phases to address their specific requirements: \textit{dense frame alignment} for training and \textit{temporal narrative guidance} for inference.

\subsection{Training Phase: Dense Frame-Level Anchoring}
\label{subsec:snp_training}

During training, we utilize the \textit{EgoVid-5M} dataset. The prompts are characterized by \textbf{dense, exhaustive descriptions} aligned with specific frame ranges. This verbosity forces the model to learn fine-grained correspondences between visual features (e.g., texture, object placement) and textual concepts.

Table~\ref{tab:snp_training} presents actual samples from the dataset (Kitchen Scene). Note the high density of object descriptions (\textit{blue tiles, stainless steel sink, utensils}) which anchors the scene's identity across different clips.

\begin{longtable}{@{}p{0.18\linewidth} p{0.78\linewidth}@{}}
    \caption{\textbf{Training SNP Examples (Real Data from EgoVid).} These prompts are strictly aligned with frame indices and contain dense scene details to mitigate forgetting.} \label{tab:snp_training} \\
    \toprule
    \textbf{Frame Range} & \textbf{Dense Training Prompt (SNP)} \\
    \midrule
    \endfirsthead
    
    \multicolumn{2}{c}%
    {{\bfseries \tablename\ \thetable{} -- continued from previous page}} \\
    \toprule
    \textbf{Frame Range} & \textbf{Dense Training Prompt (SNP)} \\
    \midrule
    \endhead
    
    \midrule
    \multicolumn{2}{r}{{Continued on next page}} \\
    \bottomrule
    \endfoot
    
    \bottomrule
    \endlastfoot
    
    \textbf{Frames 0 -- 120} \newline \textit{Overview} & 
    The video begins with a view of a \textbf{kitchen sink area}, featuring a \textbf{stainless steel sink} with two compartments. The left compartment is filled with water and various utensils, including a \textbf{green cutting board}, a knife, and some spoons. The right compartment is empty. On the countertop next to the sink, there are several items, including a \textbf{bottle of dish soap}, a sponge, a knife, and a plate with some food remnants. A hand appears in the frame, pointing towards the sink area. The hand then moves to the right side of the sink, where an \textbf{orange} is placed on a plate. The hand continues to move around the sink area, pointing at different objects. \\
    \cmidrule(l){1-2}
    
    \textbf{Frames 416 -- 536} \newline \textit{Cabinet Interaction} & 
    The video showcases a small, well-organized kitchen with \textbf{blue tiled walls} and \textbf{white cabinets}. The countertop is neatly arranged with various items, including a \textbf{bottle of dish soap}, a plate with an orange and some greens, a pot on the stove, and some utensils. A person wearing a \textbf{red shirt} is seen reaching for the dish soap bottle, opening the cabinet above the sink, and then closing it. The camera angle shifts slightly to provide a closer view of the countertop and the items on it, emphasizing the tidy and organized nature of the kitchen. \\
    \cmidrule(l){1-2}
    
    \textbf{Frames 938 -- 1058} \newline \textit{Detailed Wall View} & 
    The video begins with a close-up view of a \textbf{blue tiled wall} in a kitchen, featuring a \textbf{white cabinet} above it. The camera then pans to the right, revealing more of the kitchen counter and additional blue tiles on the wall. A white cabinet door is visible, and as the camera continues to move, it shows the interior of the cabinet, which contains various items such as a \textbf{red container} and some white dishes. The camera then moves back to the left, providing a broader view of the kitchen area, including a sink and some utensils. \\

\end{longtable}

\subsection{Inference Phase: Temporal Narrative Guidance}
\label{subsec:snp_inference}

During inference, users provide prompts defined by time segments (seconds). The SNP strategy shifts focus to \textbf{temporal progression} (e.g., lighting changes) and \textbf{attribute locking} (e.g., clothing consistency).

Table~\ref{tab:snp_inference} demonstrates a full generated sequence of a professional presenter (0s -- 60s). The prompts explicitly guide the environment from \textit{dusk} to \textit{deepest night} while ensuring the subject's attributes (\textit{blonde hair, grey blazer}) remain strictly consistent across all 12 segments.

\begin{longtable}{@{}p{0.15\linewidth} p{0.81\linewidth}@{}}
    \caption{\textbf{Inference SNP Sequence (Presenter).} A continuous 60-second generation task guided by 12 sequential prompts.} \label{tab:snp_inference} \\
    \toprule
    \textbf{Time Segment} & \textbf{Inference Prompt (SNP)} \\
    \midrule
    \endfirsthead
    
    \multicolumn{2}{c}%
    {{\bfseries \tablename\ \thetable{} -- continued from previous page}} \\
    \toprule
    \textbf{Time Segment} & \textbf{Inference Prompt (SNP)} \\
    \midrule
    \endhead
    
    \midrule
    \multicolumn{2}{r}{{Continued on next page}} \\
    \bottomrule
    \endfoot
    
    \bottomrule
    \endlastfoot
    
    \textbf{0s -- 5s} & 
    A woman with \textbf{short blonde hair}, wearing a \textbf{light grey blazer} over a white top, stands in front of a \textbf{cityscape at dusk}. She is adorned with \textbf{large, round earrings} and has a microphone clipped to her blazer. The background features tall skyscrapers with illuminated windows, reflecting the city lights. The scene emphasizes a professional, realistic style. \\
    \cmidrule(l){1-2}

    \textbf{5s -- 10s} & 
    A woman with short blonde hair, wearing a light grey blazer over a white t-shirt adorned with a small black heart, stands in front of a cityscape at dusk. She is equipped with a \textbf{black microphone} and small, round, \textbf{silver-colored earrings}. The background features skyscrapers and a \textbf{twilight sky} transitioning from blue to purple hues. She gestures while looking directly at the camera. \\
    \cmidrule(l){1-2}

    \textbf{10s -- 15s} & 
    The woman maintains a neutral expression, looking directly at the camera. Her attire---a \textbf{light grey blazer} over a white t-shirt with a microphone graphic---remains consistent. The background persists as a \textbf{cityscape at dusk} with tall skyscrapers and a twilight sky, reinforcing the professional urban environment. \\
    \cmidrule(l){1-2}

    \textbf{15s -- 20s} & 
    Standing in a professional studio setting, she holds a pen in her right hand. She is adorned with a \textbf{black, curved hair clip}. Behind her, the cityscape with tall buildings and a \textbf{twilight sky} forms the backdrop. The composition remains clean and professional, with the woman as the focal point. \\
    \cmidrule(l){1-2}

    \textbf{20s -- 25s} & 
    The woman stands in front of the cityscape at \textbf{dusk or night}. She speaks and gestures. The background features tall buildings with illuminated windows against a twilight gradient (blue to purple). The modern buildings with sleek lines contribute to a contemporary skyline as she delivers her speech. \\
    \cmidrule(l){1-2}

    \textbf{25s -- 30s} & 
    She maintains a serious expression with occasional lip movements. Her \textbf{large, circular earrings} are visible as she holds a white object. The background features skyscrapers reflecting city lights against a \textbf{twilight sky}. The blazer has a notched lapel and single-breasted design. \\
    \cmidrule(l){1-2}

    \textbf{30s -- 35s} & 
    The woman adjusts her blazer collar. Behind her, the cityscape now shows \textbf{deeper twilight hues} with the first visible stars appearing. She holds a tablet in her left hand while gesturing with her right. The microphone's LED indicator glows red, confirming active recording. \\
    \cmidrule(l){1-2}

    \textbf{35s -- 40s} & 
    With a relaxed posture, she leans against a high table. Her t-shirt's graphic aligns with her actual lapel mic. The city backdrop now features \textbf{animated data visualizations} projected onto the building facades, synchronizing with her speech about urban analytics. \\
    \cmidrule(l){1-2}

    \textbf{40s -- 45s} & 
    A close-up shot reveals precise hand gestures. The lighting has progressed to \textbf{full night}, transforming the windows of distant skyscrapers into grids of golden light. Her blazer's single button is now fastened, marking a transition segment. \\
    \cmidrule(l){1-2}

    \textbf{45s -- 50s} & 
    The scene widens to reveal a second camera operator reflected in the studio glass. The presenter's \textbf{blonde hair} shows subtle highlights under the key lights. The cityscape now includes \textbf{moving light trails} from elevated trains weaving through the urban canyon. \\
    \cmidrule(l){1-2}

    \textbf{50s -- 55s} & 
    A sudden laugh breaks her professional demeanor. In the background, a \textbf{fireworks display} erupts over the city skyline, its colored bursts reflecting in the studio's glass surfaces, adding dynamic energy to the night scene. \\
    \cmidrule(l){1-2}

    \textbf{55s -- 60s} & 
    She stands beside a transparent touchscreen displaying metropolitan statistics. The \textbf{deepest night sky} makes the illuminated buildings appear to float in darkness, their glass facades mirroring the studio setup. \\

\end{longtable}

\end{document}